\documentclass[11pt,a4paper]{article}
\usepackage{bbm}
\usepackage{amssymb}
\usepackage{amsmath}
\usepackage{graphicx}
\usepackage{subfigure}
\usepackage{cite}
\usepackage{algorithm}
\usepackage[scale={0.72,0.76}, hmarginratio=1:1, vmarginratio=1:1, headheight=2ex, headsep = 2ex, footskip = 3.6ex, nomarginpar]{geometry}
\usepackage{algorithmic}
\usepackage{color}
\newtheorem{theorem}{Theorem}

\newtheorem{lemma}{Lemma}

\newtheorem{proposition}{Proposition}

\newtheorem{assumption}{Assumption}

\begin{document}
\title{Construction of neural networks for realization of localized deep learning}

\author{Charles K. Chui$^{1,2}$ \and Shao-Bo
Lin$^3$
 \and Ding-Xuan Zhou$^4$}

  \date{}
\maketitle

\begin{center}

 \footnotesize 
1. Department of Mathematics, Hong Kong Baptist University, Kowloon,
Hong Kong, China

2. Department of Statistics,  Stanford University, CA 94305, USA

3. Department of Mathematics, Wenzhou University, Wenzhou 325035,
China

4. Department of Mathematics, City University of Hong Kong, Kowloon,
 Hong Kong, China

\begin{abstract}
The subject of deep learning has recently attracted users of machine
learning from various disciplines, including: medical diagnosis and
bioinformatics, financial market analysis and online advertisement,
speech and handwriting recognition, computer vision and natural
language processing, time series forecasting, and search engines.
However, theoretical development of deep learning is still at its
infancy. The objective of this paper is to introduce a deep neural
network (also called deep-net) approach to localized manifold
learning, with each hidden layer endowed with a specific learning
task. For the purpose of illustrations, we only focus on deep-nets
with three hidden layers, with the first layer for dimensionality
reduction, the second layer for bias reduction, and the third layer
for variance reduction. A feedback component also designed to
eliminate outliers. The main theoretical result in this paper is the
order $\mathcal O\left(m^{-2s/(2s+d)}\right)$ of approximation of
the regression function with regularity $s$, in terms of the number
$m$ of sample points, where the (unknown) manifold dimension $d$
replaces the dimension $D$ of the sampling (Euclidean) space for
shallow nets.
\end{abstract}
\end{center}

\section{Introduction}
The continually rapid growth in data acquisition and data updating
has recently posed crucial challenges to the machine learning
community on developing learning schemes to match or outperform
human learning  capability. Fortunately, the introduction of deep
learning (see, for example, \cite{Hinton2006}) has led to the
feasibility of getting around the bottleneck of classical learning
strategies, such as the support vector machine and boosting
algorithms, based on classical neural networks (see, for example,
\cite{Lippmann1987,Funahashi1989,Cybenko1989,Chui1992}), by
demonstrating remarkable successes in many applications,
particularly computer vision \cite{Krizhevsky2012} and speech
recognition \cite{Lee2009}, and more currently in other areas,
including: natural language processing, medical diagnosis and
bioinformatics, financial market analysis and online advertisement,
time series forecasting and search engines. Furthermore, the
exciting recent advances of deep learning schemes for such
applications have motivated the current interest in re-visiting the
development of classical neural networks (to be called ''shallow
nets'' in later discussions), by allowing multiple hidden layers
between the input and output layers. Such neural networks are called
''deep'' neural nets, or simply, deep nets (DN). Indeed, the
advantages of DN's over shallow nets, at least in applications, have
led to various popular research directions in the academic
communities of Approximation Theory and Learning Theory. Explicit
results on the existence of functions, that are expressible by DN's
but cannot be approximated by shallow nets with comparable number of
parameters, are generally regarded as powerful features of the
advantage of DN's in Approximation Theory. The first theoretical
understanding of such results dates back to our early work
\cite{Chui1994}, where by using the Heaviside activation function,
it was shown that DN's with two hidden layers already provide
localized approximation, while shallow nets fail. Later explicit
results on DN approximation
\cite{Eldan2015,Mhaskar2016,Telgarsky2016,Raghu2016,Poggio2017}
further reveal other various advantages of DN's over shallow nets.

From approximation to learning, the tug of war between bias and
variance \cite{Cucker2007} indicates that explicit derivation of
DN's is insufficient to show its success in machine learning, in
that besides bias, the capacity of DN should possess the
expressivity of embodying variance. In this direction, the capacity
of DN's, as measured by the number of linear regions, Betty number,
neuron transitions, and DN trajectory length were studied in
\cite{Montufar2013}, \cite{Bianchini2014}, and \cite{Raghu2016}
respectively, in showing that DN's allow for many more
functionalities than shallow nets. Although these results certainly
show the benefits of deep nets, yet they pose more difficulties in
analyzing the deep learning performance, since large capacity
usually implies large variance and requires more elaborate learning
algorithms. One of the main difficulties is development of
satisfactory learning rate analysis for DN learning, that has been
well studied for shallow nets (see, for example,
\cite{Maiorov2006a}). In this paper, we present an analysis of the
advantages of DN's in the framework of learning theory
\cite{Cucker2007}, taking into account the trade-off between bias
and variance.

Our starting point is to assume that the samples are located
approximately on some unknown manifold in the sample
($D$-dimensional Euclidean) space. For simplicity, consider the set
of {  inputs of samples}: $ x_1, \dots, x_m \in\mathcal
X\subseteq[-1,1]^D$, with a corresponding set of {  outputs}: $ y_1,
\cdots, y_m \in\mathcal Y\subseteq [-M,M]$ for some positive number
$M$, where $\mathcal X$ is an unknown data-dependent $d$-dimensional
connected $C^\infty$ Riemannian manifold (without boundary). We will
call $S_m=\{(x_i,y_i)\}_{i=1}^m$ the sample set, and construct a DN
with three hidden layers, with the first for the
dimensionality-reduction, the second for bias-reduction, and the
third for variance-reduction. The main tools for our construction
are the ``local manifold learning'' for deep nets in
\cite{Chui2016}, ``localized approximation'' for deep nets in
\cite{Chui1994}, and ``local average'' in \cite{Gyorfi2002}. We will
also introduce a  feedback procedure to eliminate outliers during
the learning process.  Our constructions justify the common
consensus that deep nets are intuitively capable of capturing data
features via their architectural structures \cite{Bengio2009}. In
addition, we will prove that the constructed DN can well approximate
the so-called regression function \cite{Cucker2007} within the
accuracy of $\mathcal O\left(m^{-2s/(2s+d)}\right)$ in expectation,
where $s$ denotes the order of smoothness (or regularity) of the
regression function. Noting that the best existing learning rates of
the shallow nets are $\mathcal O\left(m^{-2s/(2s+D)}\log^2m\right)$
\cite{Maiorov2006a} and $\mathcal O\left(m^{-s/(8s+4d)}(\log
m)^{s/(4s+2d)}\right)$ \cite{Ye2008}, we observe the power of deep
nets over shallow nets, at least theoretically, in the framework of
Learning Theory.

The organization of this paper is as follows. In the next section,
we present a detailed construction of the proposed deep net. The
main results of the paper will be stated in Section
\ref{Sec.learning rate}, where { tight} learning rates of the
constructed deep net are also deduced. Discussions of our
contributions along with comparison with some related work and
proofs of the main results will be presented in Section
\ref{Sec.Comparison} and \ref{Sec.Proof1}, respectively.

%

\section{Construction of Deep Nets}\label{Sec.Construction}
In this section, we present a construction of deep neural networks
(called deep nets, for simplicity) with three hidden layers to
realize certain deep learning algorithms, by applying the
mathematical tools of localized approximation in \cite{Chui1994},
local manifold learning in \cite{Chui2016}, and local average
arguments in \cite{Gyorfi2002}. Throughout this paper, we will
consider only two activation functions: the Heaviside
function$\sigma_0$  and  the square-rectifier $\sigma_2$, where the
standard notation $t_{+}=\max\{0,t\}$ is used to define $
\sigma_n(t)=t_{+}^n= (t_{+})^n$, for any non-negative integer $n$.

\subsection{Localized approximation and localized manifold learning}\label{Subsec:Localized approximation}
Performance comparison between deep nets and shallow nets is a
classical topic in Approximation Theory.  It is well-known from
numerous publications (see, for example,
\cite{Chui1994,Eldan2015,Raghu2016,Telgarsky2016}) that various
functions can be well approximated by deep nets but not by any
shallow net with the same order of magnitude in the numbers of
neurons. In particular, it was proved in \cite{Chui1994} that deep
nets can provide localized approximation, while shallow nets fail.

For $r,q\in\mathbb N$ and an arbitrary ${\bf j}\in\mathbb N_{2q}^r$,
where $\mathbb N_{2q}^r=\{1,2,\dots,2q\}^r$, let $\zeta_{\bf{j}}
=\zeta_{{\bf j}, q} =(\zeta_{\bf j}^{(\ell)})_{\ell=1}^r
\in(-1,1)^r$ with $\zeta_{\bf j}^{(\ell)}= -1+\frac{2{\bf
j}^{(\ell)}-1}{2q} \in (-1,1)$. For $a>0$ and $\zeta\in\mathbb R^r$,
let us denote by $A_{r,a, \zeta} =\zeta + \left[-\frac{a}{2},
\frac{a}{2}\right]^r$, the cube in $\mathbb R^r$ with center $\zeta$
and width $a$. Furthermore, we define $N_{1,r, q, \zeta_{\bf j}}:
{\mathbb R}^r \to \mathbb R$ by
\begin{equation}\label{NN for localization}
      N_{1,r, q, \zeta_{\bf j}}(\xi)
      =
      \sigma_0\left\{\sum_{\ell=1}^r\sigma_0\left[\frac1{2q}+\xi^{(\ell)}-\zeta_{\bf j}^{(\ell)}\right]
      +\sum_{\ell=1}^r\sigma_0\left[\frac1{2q}-\xi^{(\ell)}+\zeta_{\bf j}^{(\ell)}\right]- 2r+\frac12 \right\}.
\end{equation}
In what follows, the standard notion $I_A$ of the indicator function
of a set (or an event) $A$ will be used. For $x\in \mathbb R$, since
\begin{eqnarray*}
     \sigma_0 \left[\frac1{2q}+x\right] + \sigma_0 \left[\frac1{2q}
    - x\right] -2 &=& I_{[-1/(2q), \infty)} (x) + I_{(-\infty,
    1/(2q)]}(x)-2\\
 &=&
\left\{\begin{array}{ll} 0, & \hbox{if} \ x\in [-1/(2q), 1/(2q)], \\
-1, & \hbox{otherwise},
\end{array}\right.
\end{eqnarray*}
 we observe that
$$ \sum_{\ell=1}^r\sigma_0\left[\frac1{2q}+\xi^{(\ell)}\right]
      +\sum_{\ell=1}^r\sigma_0\left[\frac1{2q}-\xi^{(\ell)}\right]- 2r+\frac12 \ \left\{\begin{array}{ll}
      =\frac{1}{2},& \mbox{
for}\  x \in [-1/(2q), 1/(2q)]^r,\\
 \leq-\frac{1}{2},& \mbox{otherwise.}\end{array}\right.
$$
This implies that $N_{1,r, q, \zeta_{\bf j}}$ as introduced in
(\ref{NN for localization}), is the indicator function of the cube
$\zeta_{\bf j} + [-1/(2q), 1/(2q)]^r = A_{r, 1/q, \zeta_{\bf j}}$.
Thus, the following proposition which describes the localized
approximation  property of $N_{1,r,q, \zeta_{\bf j}}$, can be easily
deduced by applying Theorem 2.3 in \cite{Chui1994 }.

\begin{proposition}\label{Proposition:localization}
Let $r,q\in\mathbb N$ be arbitrarily given. Then $N_{1,r, q,
\zeta_{\bf j}} ={I}_{A_{r,1/q,\zeta_{\bf j}}}$ for all ${\bf
j}\in\mathbb N_{2q}^r$.
\end{proposition}

On the other hand, it was proposed in  \cite{DiCarlo2007,Basri2016}
with practical arguments, that deep nets can tackle data in
highly-curved manifolds, while any shallow net fails. These
arguments were theoretically verified in \cite{Shaham2015,Chui2016},
with the implication that adding hidden layers to shallow nets
should enable the neural networks to have the capability of
processing massive data in a high-dimensional space from samples in
lower dimensional manifolds. More precisely, it follows from
\cite{Docarmo1992,Shaham2015} that for a lower $d$-dimensional
connected and compact $C^\infty$ Riemannian submanifold  $\mathcal X
\subseteq[-1,1]^D$ (without boundary),  isometrically embedded in
${\mathbb R}^D$  and endowed with the geodesic distance $d_G$, there
exists some $\delta>0$, such that for any $x, x'\in\mathcal X$, with
$d_G(x,x')<\delta$,
\begin{equation}\label{smooth manifold}
    \frac12d_G(x, x')\leq\|x-x'\|_D\leq2d_G(x,x'),
\end{equation}
where for any $r>0$, $\|\cdot\|_r$ denotes, as usual, the Euclidean
norm of $\mathbb R^r$. In the following, let $B_G(\xi_0,\tau)$,
$B_D(\xi_0,\tau)$, and $B_{d}(\xi_0, \tau)$ denote the closed
geodesic ball, the $D$-dimensional Euclidean ball, and the
$d$-dimensional Euclidean ball, with center  at $\xi_0$,
respectively, and with radius $\tau>0$. Then the following
proposition is a brief summary of Theorem 2.2, Theorem 2.3 and
Remark 2.1 in \cite{Chui2016}, with the implication that neural
network can be used as a dimensionality-reduction tool.

\begin{proposition}\label{Proposition:local manifold learning}
For each  $\xi\in\mathcal X$,  there exist a positive number
$\delta_\xi$ and a neural network
$$ \Phi_\xi =(\Phi^{(\ell)}_\xi)_{\ell=1}^d: {\mathcal X} \to {\mathbb R}^d$$
with
\begin{equation}\label{representation for phi x}
      \Phi_\xi^{(\ell)}(x) =\sum_{k=1}^{(D+2)(D+1)}a_{k,\xi,\ell}
      \sigma_2(w_{k,\xi,\ell}\cdot x+b_{k,\xi,\ell}), \qquad w_{k,\xi,\ell}\in\mathbb R^D, a_{k,\xi,\ell}, b_{k,\xi,\ell}\in\mathbb R,
\end{equation}
that maps $B_G(\xi,\delta_\xi)$ diffeomorphically onto $[-1,1]^d$
and satisfies
\begin{equation}\label{equality of distance}
       \alpha_\xi d_G(x,x')
       \leq
       \|\Phi_\xi(x)-\Phi_\xi(x')\|_d\leq
       \beta_\xi d_G(x,x'), \qquad  \forall\ x,x'\in
       B_G(\xi,\delta_\xi)
\end{equation}
for some $\alpha_\xi,\beta_\xi>0$.
\end{proposition}

\subsection{Learning via deep nets}\label{Subsec: learning without feedback}

Our construction of deep nets depends on the localized approximation
and dimensionality-reduction technique, as presented in Propositions
\ref{Proposition:localization} and \ref{Proposition:local manifold
learning}. To describe the learning process, firstly select a
suitable $q^*$, so that for every ${\bf j}\in N_{2q^*}^d$, there
exists some point $\xi^*_{\bf j}$ in a finite set
$\{\xi^*_{i}\}_{i=1}^{F_\mathcal X} \subset \mathcal X$ that
satisfies
\begin{equation}\label{target 1}
       A_{D,1/q^*,\zeta_{{\bf j},q^*}}\cap\mathcal X\subset B_G(\xi^*_{\bf j}, \delta_{\xi^*_{\bf j}}).
\end{equation}
To this end, we need a constant $C_0 \geq 1$, such that
\begin{equation}\label{embedRD}
d_G(x, x') \leq C_0 \|x-x'\|_D, \qquad \forall\ x,x'\in \mathcal X.
\end{equation}
The existence of such a constant is proved in the literature (see,
for example, \cite{Ye2008}). Also, in view of the compactness of
$\mathcal X$, since  $\bigcup_{\xi\in\mathcal X}\{x\in \mathcal X:
B_G(x, \xi) <\delta_\xi/2\}$ is an open covering of $\mathcal X$,
there exists a finite set of points $\{\xi^*_{i}\}_{i=1}^{F_\mathcal
X} \subset \mathcal X$, such that $\mathcal
X\subset\bigcup_{i=1}^{F_{\mathcal
X}}B_G(\xi^*_i,\delta_{\xi^*_i}/2).$ Hence, $q^* \in \mathbb N$ may
be  chosen to satisfy
\begin{equation}\label{choose q}
        q^* \geq \frac{2 C_0 \sqrt{D}}{\min_{1\leq i \leq\mathcal F_{\mathcal X}}\delta_{\xi^*_i}}.
\end{equation}
With this choice, we claim that (\ref{target 1}) holds. Indeed, if
$A_{D,1/q^*,\zeta_{{\bf j},q^*}}\cap\mathcal X=\varnothing$, then
(\ref{target 1}) obviously holds for any choice of $\xi\in\mathcal
X$. On the other hand, if $A_{D,1/q^*,\zeta_{{\bf
j},q^*}}\cap\mathcal X \neq\varnothing$, then from the inclusion
property $\mathcal X\subset\bigcup_{i=1}^{F_{\mathcal
X}}B_G(\xi^*_i,\delta_{\xi^*_i}/2)$, it follows that there is some
$i^* \in \{1, \ldots, F_\mathcal X\}$, depending on ${\bf j} \in
N_{2q^*}^d$, such that
\begin{equation}\label{tool 1 for manifold}
        A_{D,1/q^*,\zeta_{{\bf j},q^*}}\cap
        B_G(\xi^*_{i^*},\delta_{\xi^*_{i^*}}/2)\neq \varnothing.
\end{equation}
Next, let $\eta^*\in A_{D,1/q^*,\zeta_{{\bf j},q^*}}\cap
        B_G(\xi^*_{i^*},\delta_{\xi^*_{i^*}}/2)$. By (\ref{embedRD}), we have, for any $x\in
A_{D,1/q^*,\zeta_{{\bf j},q^*}}\cap\mathcal X$,  $$d_G (x, \eta^*)
\leq C_0 \|x-\eta^*\|_D \leq C_0 \sqrt{D} \frac{1}{q^*}.$$
Therefore, it follows from (\ref{choose q}) that
\begin{eqnarray*}
      d_G(x,\xi^*_{i^*})
      &\leq&
      d_G(x,\eta^*)+d_G(\eta^*,\xi_{i^*}^*)
      \leq
      C_0 \sqrt{D} \frac{1}{q^*} +\frac{\delta_{\xi_{i^*}^*}}{2}
      \leq \delta_{\xi_{i^*}^*}.
\end{eqnarray*}
This implies that $A_{D,1/q^*,\zeta_{{\bf j},q^*}}\cap\mathcal
X\subset
   B_G(\xi_{i^*}^*,\delta_{\xi_{i^*}^*})$ and verifies our claim (\ref{target 1}) with the choice of $\xi^*_{\bf j} = \xi_{i^*}^*$.

Observe that for every ${\bf j}\in\mathbb N_{2q^*}^D$ we may choose
the point $\xi^*_{\bf j}\in \mathcal X$ to define $N_{2,\bf j}=
(N_{2,\bf j}^{(\ell)})_{\ell=1}^d: {\mathcal X} \to {\mathbb R}^d$
by setting
\begin{equation}\label{N2(x)}
     N_{2,\bf j}^{(\ell)}(x):= \Phi_{\xi^*_{\bf j}}^{(\ell)} (x) =
     \sum_{k=1}^{(D+2)(D+1)}a_{k,\xi^*_{\bf j},\ell}
      \sigma_2\left(w_{k,\xi^*_{\bf j},\ell}\cdot
      x  +b_{k,\xi^*_{\bf j},\ell}\right), \qquad \ell =1, \ldots, d
\end{equation}
and apply Proposition \ref{Proposition:local manifold learning},
(\ref{target 1}),  and (\ref{representation for phi x}) to obtain
the following.

\begin{proposition}\label{Proposition:Manifold learning}
For each ${\bf j}\in\mathbb N_{2q^*}^D$, $N_{2, \bf j}$ maps
$A_{D,1/q^*,\zeta_{{\bf j},q^*}}\cap\mathcal X$ diffeomorphically
into
  $[-1,1]^d$ and
\begin{equation}\label{equality of distance of N2}
       \alpha d_G(x,x')
       \leq
       \|N_{2,\bf j}(x)-N_{2,\bf j}(x')\|_d\leq
       \beta d_G(x,x'),\qquad\forall\ x,x'\in  A_{D,1/q^*,\zeta_{{\bf j},q^*}}\cap\mathcal X,
\end{equation}
where $\alpha:=\min_{1\leq i \leq F_{\mathcal X}}\alpha_{\xi^*_{i}}$
and $\beta:=\max_{1\leq i\leq F_{\mathcal X}}\beta_{\xi^*_{i}}$.
\end{proposition}

As a result of Propositions \ref{Proposition:localization} and
\ref{Proposition:Manifold learning}, we now present the construction
of the deep nets for the proposed learning purpose. Start with
selecting $(2n)^d$ points $t_{\bf k}= t_{{\bf k}, n}\in (-1,1)^d$,
${\bf k}\in\mathbb N_{2n}^d$ and $n\in\mathbb N$, with $t_{\bf k} =
(t_{\bf k}^{1}, \cdots, t_{\bf k}^{d})$, where $t_{\bf
k}^{(\ell)}=-1+\frac{2{\bf k}^{(\ell)}-1}{2n}$ in $(-1,1)^d$. Denote
$C_{\bf k} =A_{d,1/n,t_{{\bf k}}}$ and $H_{{\bf k,j}}
=\{x\in\mathcal X\cap A_{D,1/q^*,\zeta_{{\bf j},q^*}}: N_{2,{\bf
j}}(x)\in C_{\bf k}\}$. In view of Proposition
\ref{Proposition:Manifold learning}, it follows that $H_{\bf k, j}$
is well defined, ${\mathcal X} \subseteq  \cup_{{\bf j}\in\mathbb
N_{2q^*}^D} A_{D,1/q^*,\zeta_{{\bf j},q^*}}$, and  $ \bigcup_{{\bf
k}\in\mathbb N_{2n}^d} H_{\bf k, j}
 =\mathcal X\cap A_{D,1/q^*,\zeta_{{\bf j},q^*}}.$  We also define $N_{3,{\bf k,j}}: {\mathcal X} \to {\mathbb R}$ by
\begin{eqnarray}\label{N3k(x)}
      &&N_{3,{\bf k,j}}(x) =N_{1, d, n, t_{\bf k}} \circ N_{2,{\bf j}}(x) \\
      && = \sigma_0\left\{
       \sum_{\ell=1}^d\sigma_0
           \left[\frac{1}{2n}+N^{(\ell)}_{2,{\bf j}}(x)-t_{\bf
           k}^{(\ell)}\right]
            +
            \sum_{\ell=1}^d\sigma_0
           \left[\frac1{2n}-N^{(\ell)}_{2,{\bf j}}(x)+t_{\bf
           k}^{(\ell)}\right]-2d+\frac12\right\}. \nonumber
\end{eqnarray}
Then  the desired deep net estimator with three hidden layers may be
defined by
\begin{eqnarray}\label{final estimator1}
            N_{3}(x)= \frac{\sum_{{\bf j}\in \mathbb N_{2 q^*}^D} \sum_{{\bf k}\in \mathbb N_{2n}^d} \sum_{i=1}^m N_{1,D, q^*, \zeta_{{\bf j}}} (x_i) N_{3,{\bf k,j}}(x_i) y_i N_{3,{\bf k,j}}(x)}{
           \sum_{{\bf j}\in \mathbb N_{2 q^*}^D} \sum_{{\bf k}\in \mathbb N_{2n}^d} \sum_{i=1}^m N_{1,D, q^*, \zeta_{{\bf j}}} (x_i) N_{3,{\bf k,j}}(x_i)},
\end{eqnarray}
where we set $N_{3}(x)=0$ if the denominator is zero.

Observe that in the above construction there is a totality of three
hidden layers to perform three separate tasks, namely: the task of
the first hidden layer is to reduce the dimension of the input
space, while the second and third hidden layers are to perform
localized approximation on $\mathbb R^d$ and data variance reduction
by applying local averaging \cite{Gyorfi2002}, respectively.

\subsection{Fine-tuning}
For each $x\in {\mathcal X}$, it follows from  $\mathcal
X=\bigcup_{{\bf j} \in \mathbb N_{2q^*}^D} A_{D,1/q^*,\zeta_{{\bf
j},q^*}}$  that there is some ${\bf j}\in\mathbb N_{2q^*}^D$, such
that $x\in A_{D,1/q^*,\zeta_{{\bf j},q^*}}$, which implies that
$N_{2,{\bf j}}(x) \in [-1, 1]^d$. For each ${\bf j}\in\mathbb
N_{2q}^{*}$, since  $ A_{D,1/q^*,\zeta_{{\bf j},q^*}}$ is a cube in
$\mathbb R^D$, the cardinality of the set $\{{\bf j}:x\in
A_{D,1/q^*,\zeta_{{\bf j},q^*}}\}$  is at most $2^D$. Also, because
$[-1,1]^d=\bigcup_{{\bf k} \in \mathbb N_{2n}^d}A_{d, 1/n, t_{\bf
k}}$ for each ${\bf j}\in\mathbb N_{2q}^{*}$, there exists some
${\bf k}\in \mathbb N_{2n}^d$, such that $N_{2,{\bf j}}(x) \in A_{d,
1/n, t_{\bf k}}$, implying that $N_{3,{\bf k,j}}(x) =N_{1, d, n,
t_{\bf k}}\circ N_{2,{\bf j}}(x) = 1$ and that the number of such
integers ${\bf k}$ is bounded by $2^d$. For each $x\in\mathcal X$,
we consider a non-empty subset
\begin{equation}\label{Lambdaset}
     \Lambda_x =\left\{({\bf j,k})\in\mathbb N_{2q^*}^D \times \mathbb
       N_{2n}^d: x\in A_{D,1/q^*,\zeta_{{\bf j},q^*}}, N_{3,{\bf k,j}}(x)=1 \right\}.
\end{equation}
of $\mathbb N_{2q^*}^D \times \mathbb N_{2n}^d$, with cardinality
\begin{equation}\label{cap1}
        |\Lambda_x|\leq 2^{D+d},\qquad \forall\ x\in\mathcal X.
\end{equation}
Also, for each $x\in\mathcal X$, we further define
$S_{\Lambda_x}=\cup_{({\bf j,k})\in \Lambda_x} H_{\bf k,j}\cap
\{x_i\}_{i=1}^m$, as well as
\begin{equation}\label{Lambdaset1}
     \Lambda_{x,S}=\left\{({\bf j,k})\in\mathbb
     N_{2q^*}^D\times\mathbb N_{2n}^d, N_{1,D, q^*, \zeta_{{\bf j}}} (x_i) N_{3,{\bf
     k,j}}(x_i)=1,x_i\in S_{\Lambda_x}\right\},
\end{equation}
and
\begin{equation}\label{Lambdaset2}
     \Lambda'_{x,S}=\left\{({\bf j,k})\in\mathbb
     N_{2q^*}^D\times\mathbb N_{2n}^d, N_{1,D, q^*, \zeta_{{\bf j}}} (x_i) N_{3,{\bf
     k,j}}(x_i)N_{3,{\bf
     k,j}}(x)=1,x_i\in S_{\Lambda_x}\right\}.
\end{equation}
Then it follows from  (\ref{Lambdaset1}) and (\ref{Lambdaset2}) that
$|\Lambda'_{x,S}|\leq|\Lambda_{x,S}|,$ and it is easy to see that if
each $x_i\in S_{\Lambda_x}$ is an interior point of some $H_{\bf
k,j}$, then $|\Lambda_{x,S}|=|\Lambda'_{x,S}|$. In this way, $N_3$
is some local average estimator. However, if $|\Lambda_{x,S}|\neq
|\Lambda'_{x,S}|$, (and this is possible when some $x_i$ lies on the
boundary of $H_{\bf k,j}$ for some $({\bf j,k})\in \mathbb
N_{2q^*}^D\times\mathbb N_{2n}^d$), then the estimator $N_3$
(\ref{final estimator1}) might perform badly, and this happens even
for training data. Note that to predict some $x_j\in S_m$, which is
an interior point of $H_{{\bf k}_0,{\bf j}_0}$, we have
$$
        N_3(x_j)=\frac{\sum_{i=1}^mN_{1,D,q^*,\zeta_{{\bf
        j}_0}}(x_i)N_{3,{\bf k}_0,{\bf
        j}_0}(x_i)y_i}{|\Lambda'_{x_j,S}|},
$$
which is much smaller than $y_j$ when
$|\Lambda'_{x,S}|<|\Lambda_{x,S}|$. The reason is that there are
only $|\Lambda_{x,S}|$ summations in the numerator. Noting that the
Riemannian measure of the boundary of $\cup_{({\bf j,k})\in\mathbb
N_{2q^*}^D\times\mathbb N_{2n}^d}H_{\bf k,j}$ is zero, we consider
the above phenomenon as outliers.

Fine-tuning, often referred to as feedback in the literature of deep
learning \cite{Bengio2009}, can essentially improve the learning
performance of deep nets \cite{Larochelle2009}.  We observe that
fine-tuning can also be applied to avoid outliers for our
constructed deep net in (\ref{final estimator1}), by counting the
cardinalities of $\Lambda_{x,S}$ and $\Lambda'_{x,S}$. In the
training processing, besides computing $N_3(x)$ for some query point
$x$, we may also record $|\Lambda_{x,S}|$ and $|\Lambda'_{x,S}|$. If
the estimator is not big enough, we propose to add the factor
$\frac{|\Lambda'_{x,S}|} {|\Lambda_{x,S}|}$ to $N_3(x)$. In this
way, the deep net estimator with feedback can be mathematically
represented by
\begin{equation}\label{feedback 1}
           N_3^F(x)=\frac{|\Lambda'_{x,S}|}{|\Lambda_{x,S}|}N_3(x)=\frac{\sum_{{\bf j}\in \mathbb N_{2 q^*}^D}
           \sum_{{\bf k}\in \mathbb N_{2n}^d} \sum_{i=1}^m y_i \Phi_{\bf k,j}(x, x_i)}{
           \sum_{{\bf j}\in \mathbb N_{2 q^*}^D} \sum_{{\bf k}\in
           \mathbb N_{2n}^d} \sum_{i=1}^m \Phi_{\bf k,j}(x, x_i)},
\end{equation}
where $\Phi_{\bf k,j} = \Phi_{{\bf k,j}, D, q^*, n}: {\mathcal
X}\times {\mathcal X} \to {\mathbb R}$ is defined by
$$ \Phi_{\bf k,j}(x, u) = N_{1,D, q^*, \zeta_{{\bf j}}} (u) N_{3,{\bf k,j}}(u) N_{3,{\bf k,j}}(x); $$
and as before,  we set $N_3^F(x)=0$ if the denominator $\sum_{{\bf
j}\in \mathbb N_{2 q^*}^D} \sum_{{\bf k}\in \mathbb N_{2n}^d}
\sum_{i=1}^m \Phi_{\bf k,j}(x, x_i)$ vanishes.

\section{Learning Rate Analysis}\label{Sec.learning rate}

We consider a standard regression setting in learning theory
\cite{Cucker2007} and assume that the sample set $S=
S_m=\{(x_i,y_i)\}_{i=1}^m$ of size $m$ is drawn independently
according to some Borel probability measure $\rho$ on ${\mathcal Z}
= {\mathcal X}\times {\mathcal Y}$. The regression function is then
defined by
$$
         f_\rho(x)=\int_{\mathcal Y} y d\rho(y|x), \qquad x\in\mathcal X,
$$
where $\rho(y|x)$ denotes the conditional distribution at $x$
induced by $\rho$. Let $\rho_X$ be the marginal distribution of
$\rho$ on $\mathcal X$ and $(L^2_{\rho_{_X}}, \|\cdot\|_\rho)$ be
the Hilbert space of square-integrable functions with respect to
$\rho_X$ on $\mathcal X$. Our goal is to estimate the distance
between the output function $N_3$  and the regression function
$f_\rho$ measured by $\|N_3-f_\rho\|_\rho$, as well as the distance
between $N_3^F$  and $\|N_3^F-f_\rho\|_\rho$.

We say that a function $f$ on $\mathcal X$ is $(s,c_0)$-Lipschitz
(continuous) with positive exponent $s \leq 1$ and constant $c_0
>0$, if
\begin{equation}\label{Smoothness assumption}
     |f(x)-f(x')|\leq c_0(d_G(x,x'))^s, \qquad \forall   x,x'\in\mathcal X;
\end{equation}
and denote by $Lip^{(s,c_0)}:=Lip^{(s,c_0)}(\mathcal X)$, the family
of all $(s,c_0)$ Lipschitz functions that satisfy (\ref{Smoothness
assumption}). Our error analysis of $N_3$ will be carried out based
on the following two assumptions.

\begin{assumption}\label{Assumption:frho}
There exist an $s\in(0,1]$ and a constant $c_0\in\mathbb R_+$ such
that $f_\rho\in Lip^{(s,c_0)}$.
\end{assumption}

This smoothness assumption is standard in learning theory for the
study of approximation for regression (see, for example,
and\cite{Gyorfi2002,Kohler2005,Maiorov2006a,Cucker2007,Wu2008,Shi2011,Hu2015,Fan2016,Guo2016,Christmann2016,Chang2017,Lin2017}).
\begin{assumption}\label{Assumption:rhox}
  $\rho_X$ is continuous with respect to the geodesic distance $d_G$ of the Riemannian manifold.
\end{assumption}

Note that Assumption \ref{Assumption:rhox}, which is about the
geometrical structure of $\rho_X$, is slightly weaker than the
distortion assumption in \cite{Zhou2006,Shi2013} but somewhat
similar to the assumption considered in \cite{Meister2016}. The
objective of this assumption is for describing the functionality of
fine-tuning.

We are now ready to state the main results of this paper. In the
first theorem below, we obtained an upper bound of learning rate for
the constructed deep nets $N_3$.

\begin{theorem}\label{Theorem: optimal rate without feedback}
Let $m$ be the number of samples and set $n=\lceil
m^{1/(2s+d)}\rceil$, where $1/(2n)$ is the uniform spacing of the
points $t_{\bf k}= t_{{\bf k}, n}\in (-1,1)^d$ in the definition of
$ N_3$ in (\ref{N3k(x)}). Then under Assumptions
\ref{Assumption:frho} and \ref{Assumption:rhox},
\begin{equation}\label{theorem1}
         \mathbf
        E\left[\|N_3-f_\rho\|_\rho^2\right]
        \leq
        C_1m^{-\frac{2s}{2s+d}}.
\end{equation}
for some positive constant $C_1$ independent of $m$.
\end{theorem}

Observe that Theorem \ref{Theorem: optimal rate without feedback}
provides fast learning rate for the constructed deep net  which
depends on manifold dimension $d$ instead of the sample space
dimension $D$. In the second theorem below, we show the necessity of
the fine-tuning process as presented in (\ref{feedback 1}), when
Assumption \ref{Assumption:rhox} is removed.
\begin{theorem}\label{Theorem: optimal rate}
Let $m$ be the number of samples and set $n=\lceil
m^{1/(2s+d)}\rceil$, where $1/(2n)$ is the uniform spacing of the
points $t_{\bf k}= t_{{\bf k}, n}\in (-1,1)^d$ in the definition of
$N_3$ in (\ref{N3k(x)}), which is used to define $N_3^F$ in
(\ref{feedback 1}). Then under Assumption \ref{Assumption:frho},
\begin{equation}\label{theorem2}
         \mathbf
        E\left[\|N_3^F-f_\rho\|_\rho^2\right]
        \leq
        C_2'm^{-\frac{2s}{2s+d}}.
\end{equation}
for some positive constant $C_2$ independent of $m$.
\end{theorem}

Observe that while Assumption \ref{Assumption:rhox}  is needed in
Theorem \ref{Theorem: optimal rate without feedback}, it is not
necessary for the validity of Theorem \ref{Theorem: optimal rate},
which theoretically shows the significance of fine-tuning in our
construction. The proofs of these two theorems will be presented in
the final section of this paper.

\section{Related Work and Discussions}\label{Sec.Comparison}

The success in practical applications,  especially in the fields of
computer vision \cite{Krizhevsky2012} and speech recognition
\cite{Lee2009},  has triggered enormous research activities on deep
learning. Several other encouraging results, such as object
recognition \cite{DiCarlo2007}, unsupervised training
\cite{Erhan2010}, and artificial intelligence
architecture\cite{Bengio2009}, have been obtained to demonstrate the
significance of deep learning. We refer the interested readers to
the 2016 MIT monograph, ``Deep Learning'' \cite{Goodfellow}, by
Goodfellow, Bengjio and Courville, for further study of this
exciting subject, which is only at the infancy of its development.

Indeed, deep learning has already created several challenges to the
machine learning community. Among the main challenges are to show
the necessity of the usage of deep nets and to theoretically justify
the advantages of deep nets over shallow nets. This is essentially a
classical topic in Approximation Theory. In particular, dating back
to the early 1990's, it was already proved that deep nets can
provide localized approximation but shallow nets fail (see, for
example, \cite{Chui1994}). Furthermore, it was also shown that deep
nets provide high approximation orders, that are certainly not
restricted by the lower error bounds for shallow nets (see
\cite{Chui1996, Maiorov1999b}). More recently, stimulated by the
avid enthusiasm of deep learning, numerous advantages of deep nets
were also revealed from the point of view of function approximation.
In particular, certain functions discussed in \cite{Eldan2015} can
be represented by deep nets but cannot be approximated by shallow
nets; it was shown in \cite{Mhaskar2016} that deep nets, but not
shallow nets, can approximate composition of functions; it was
exhibited in\cite{Poggio2017} that deep nets can  avoid the curse of
dimension of shallow nets; a probability argument was given in
\cite{Lin2017a} to show that deep nets have better approximation
performance than shallow nets with high confidence; it was
demonstrated in \cite{Shaham2015,Chui2016} that deep nets can
improve the approximation capability of shallow nets when the data
are located on data-dependent manifolds; and so on.  All of these
results give theoretical explanations of the significance of deep
nets from the Approximation Theory point of view.

As a departure from the work mentioned above, our present paper is
devoted to explore better performance of deep nets over shallow nets
in the framework of Leaning Theory. In particular, we are concerned
not only with the approximation accuracy but also with the cost to
attain such accuracy. In this regard, learning rates of certain deep
nets have been analyzed in \cite{Kohler2005}, in which Kohler and
Krzy\.{z}ak provided certain near-optimal learning rates for a
fairly complex regularization scheme, with the hypothesis space
being the family of deep nets with two hidden layers proposed in
\cite{Mhaskar1993}. More precisely, they derived a learning rate of
order $\mathcal O(m^{-2s/(2s+D)}(\log m)^{4s/(2s+D)})$ for functions
$f_\rho\in Lip^{(s,c_0)}$. This is close to the optimal learning
rate of shallow nets in \cite{Maiorov2006a}, different only by a
logarithmic factor. Hence, the study in \cite{Kohler2005}
theoretically showed that deep nets at least do not downgrade the
learning performance of shallow nets. In comparison with
\cite{Kohler2005}, our study is focussed on answering the question:
''What is to be gained by deep learning?'' The deep net constructed
in our paper possesses a learning rate of  order $\mathcal
O(m^{-2s/(2s+d)})$, when $\mathcal X$ is an unknown $d$-dimensional
connected $C^\infty$ Riemannian manifold (without boundary). This
rate is the same as the optimal learning rate \cite[Chapeter
3]{Gyorfi2002} for special case of the cube $\mathcal X=[-1,1]^d$
under a similar condition, though it is smaller than the optimal
learning rates for shallow nets \cite{Maiorov2006a}. Another line of
related work is \cite{Ye2008,Ye2009}, where Ye and Zhou deduced
learning rates for regularized least-squares over shallow nets for
the same setting of our paper. They derived a learning rate of
$\mathcal O\left( m^{-s/(8s+4d)}(\log m)^{s/(4s+2d)}\right)$, which
is slower than the rate established in our paper.  It should be
mentioned that in a more recent work \cite{Kohler2017}, some
advantages of deep nets are revealed from the learning theory
viewpoint. However, the results in \cite{Kohler2017} requires a
hierarchical interaction structure, which is totally different from
what is presented in our present paper.

Due to the high degree of freedom for deep nets, the number and type
of parameters for deep nets are much more than those of shallow
nets. Thus, it should be of great interest to develop scalable
algorithms to reduce the computational burdens of deep learning.
Distributed learning based on a divide-and-conquer strategy
\cite{Zhang2014,Lin2015} could be a fruitful approach for this
purpose. It is also of interest to establish results similar to
those of Theorem \ref{Theorem: optimal rate} and Theorem
\ref{Theorem: optimal rate without feedback} for deep nets, but with
rectifier neurons, by using the rectifier (or ramp) function, $
\sigma_2(t)=t_{+}^2= (t_{+})^2$, as activation. The reason is that
the rectifier is one of the most widely used activations in the
literature on deep learning. Our research in these directions is
postponed to a later work.

\section{Proofs of the main results}\label{Sec.Proof1}

To facilitate our proofs of the theorems stated in Section
\ref{Sec.learning rate}, we first establish the following two
lemmas.

Observe from Proposition \ref{Proposition:localization} and the
definition (\ref{N3k(x)}) of the function $N_{3,{\bf k,j}}$ that
\begin{equation}\label{N1&3}
 N_{1,D, q^*, \zeta_{{\bf j}}} (x) N_{3,{\bf k,j}}(x) = {I}_{A_{D,1/q^*,\zeta_{\bf
j}}} (x) I_{A_{d, 1/n, t_{\bf k}}} (N_{2, {\bf j}} (x)) = I_{H_{\bf
k, j}}(x).
\end{equation}
For ${\bf j}\in \mathbb N_{2 q^*}^D, {\bf k}\in \mathbb N_{2n}^d$,
define a random function $T_{{\bf k, j}}: {\mathcal Z}^m \to \mathbb
R$ in term of the random sample $S=\{(x_i,y_i)\}_{i=1}^m$ by
\begin{equation}\label{def. T}
           T_{{\bf k, j}}(S) = \sum_{i=1}^m N_{1,D, q^*, \zeta_{{\bf j}}} (x_i) N_{3,{\bf k,j}}(x_i),
\end{equation}
so that
\begin{equation}\label{expressT}
         T_{{\bf k, j}}(S) =\sum_{i=1}^m I_{H_{\bf k, j}}(x_i).
\end{equation}

\begin{lemma}\label{Lemma:important}
Let  $\Lambda^*\subseteq \mathbb N_{2q^*}^D\times\mathbb N_{2n}^d$
be a non-empty subset, $({\bf j}\times{\bf k}) \in \Lambda^*$ and
$T_{{\bf k, j}}(S)$ be defined as in (\ref{def. T}). Then
\begin{equation}\label{lemmafor bio}
        \mathbf E_S \left[\frac{I_{\{z\in\mathcal Z^m:\sum_{({\bf j,k})\in \Lambda^*}T_{{\bf k,
j}}(z)>0\}}(S)}{\sum_{({\bf j,k})\in\Lambda^*}T_{{\bf k,
j}}(S)}\right]\leq \frac{2}{(m+1)\rho_X(\cup_{({\bf
j,k})\in\Lambda^*}H_{\bf k, j})},
\end{equation}
where if $\sum_{({\bf j,k})\in \Lambda^*}T_{{\bf k, j}}(S)=0$, we
set
$$\frac{I_{\{z\in\mathcal Z^m:\sum_{({\bf j,k})\in
\Lambda^*}T_{{\bf k, j}}(z)>0\}}(S)}{\sum_{{\bf
j,k}\in\Lambda^*}T_{{\bf k, j}}(S)}=0.
$$
\end{lemma}

{\bf Proof} Observe that it follows from (\ref{expressT}) that
$T_{{\bf k, j}}(S)\in \{0, 1, \ldots, m\}$ and
\begin{eqnarray*}
     &&\mathbf E_S \left[\frac{I_{\{z\in\mathcal Z^m:\sum_{({\bf j,k})\in \Lambda^*}T_{{\bf k,
     j}}(z)>0\}}(S)}{\sum_{({\bf j,k})\in\Lambda^*}T_{{\bf k,
      j}}(S)}\right]\\
     &=&
     \sum_{\ell=0}^{m}
     \mathbf E_S \left[\frac{I_{\{z\in\mathcal Z^m:\sum_{({\bf j,k})\in \Lambda^*}T_{{\bf k,
j}}(z)>0\}}(S)}{\sum_{({\bf j,k})\in\Lambda^*}T_{{\bf k,
j}}(S)}\big|   \sum_{({\bf j,k})\in\Lambda^*}T_{{\bf k, j}}(S)=\ell
             \right]
             Pr\left[\sum_{({\bf j,k})\in\Lambda^*}T_{{\bf k, j}}(S)=\ell\right].
\end{eqnarray*}
Since by the definition of the fraction $\frac{I_{\{z\in\mathcal
Z^m:\sum_{({\bf j,k})\in \Lambda^*}T_{{\bf
k,j}}(z)>0\}}(S)}{\sum_{({\bf j,k})\in\Lambda^*}T_{{\bf k, j}}(S)}$,
the term with $\ell=0$ above vanishes, so that
\begin{eqnarray*}
    \mathbf E_S \left[\frac{I_{\{z\in\mathcal Z^m:\sum_{({\bf j,k})\in \Lambda^*}T_{{\bf k,
j}}(z)>0\}}(S)}{\sum_{({\bf j,k})\in\Lambda^*}T_{{\bf k,
j}}(S)}\right]
   &=&\sum_{\ell=1}^{m}\mathbf E\left[\frac{1}{\ell} \big|\sum_{({\bf j, k})\in \Lambda^*}T_{{\bf k, j}}(S)=\ell\right]
   Pr\left[\sum_{({\bf j, k})\in \Lambda^*}T_{{\bf k, j}}(S)=\ell\right]\\
   & =&
   \sum_{\ell=1}^{m} \frac{1}{\ell} Pr\left[\sum_{({\bf j, k})\in \Lambda^*}T_{{\bf k, j}}(S)=\ell\right].
\end{eqnarray*}
On the other hand, from (\ref{expressT}), note that $\sum_{({\bf
j,k})\in\Lambda^*}T_{{\bf k, j}}(S)=\ell$ is equivalent to $x_i \in
\cup_{({\bf j,k})\in\Lambda^*}H_{\bf k,j}$ for $\ell$ indices $i$
from $\{1,\cdots, m\}$, which in turn implies that
$$
       Pr\left[\sum_{({\bf j, k})\in \Lambda^*}T_{{\bf k, j}}(S)=\ell\right]
       =\left(\begin{array}{c}m\\
       \ell\end{array}\right)[\rho_X(\cup_{({\bf j,k})\in\Lambda^*}
       H_{\bf k, j})]^\ell[1-\rho_X(\cup_{({\bf j,k})\in\Lambda^*}H_{\bf
       k, j})]^{m-\ell}.
$$
Thus, we obtain
\begin{eqnarray*}
      &&\mathbf E_S \left[\frac{I_{\{z\in\mathcal Z^m:\sum_{({\bf j,k})\in \Lambda^*}T_{{\bf k,
j}}(z)>0\}}(S)}{\sum_{({\bf j,k})\in\Lambda^*}T_{{\bf k,
j}}(S)}\right]\\
   & =&  \sum_{\ell=1}^{m} \frac{1}{\ell} \left(\begin{array}{c}m\\
       \ell\end{array}\right)[\rho_X(\cup_{({\bf j,k})\in\Lambda^*}
       H_{\bf k, j})]^\ell[1-\rho_X(\cup_{({\bf j,k})\in\Lambda^*}H_{\bf
       k, j})]^{m-\ell}\\
      &\leq&
      \sum_{\ell=1}^{m} \frac{2}{\ell+1} \left(\begin{array}{c}m\\
       \ell\end{array}\right)[\rho_X(\cup_{({\bf j,k})\in\Lambda^*}
       H_{\bf k, j})]^\ell[1-\rho_X(\cup_{({\bf j,k})\in\Lambda^*}H_{\bf
       k, j})]^{m-\ell}\\
        &=& \frac{2}{(m+1)\rho_X(\cup_{({\bf j,k})\in\Lambda^*}
       H_{\bf k, j}))}\sum_{\ell=1}^{m}
       \left(\begin{array}{c}m+1\\
       \ell+1\end{array}\right)[\rho_X(\cup_{({\bf j,k})\in\Lambda^*}
       H_{\bf k, j})]^{\ell+1}[1-\rho_X({\cup}_{({\bf j,k})\in\Lambda^*}
       H_{\bf k, j})]^{m-\ell}.
\end{eqnarray*}
Therefore, the desired inequality (\ref{lemmafor bio}) follows. This
completes the proof of Lemma \ref{Lemma:important}. $\Box$

\begin{lemma}\label{Lemma:important2}
Let $S=\{(x_i,y_i)\}_{i=1}^m$ be a sample set drawn independently
according to $\rho$. If $f_{S}(x)=\sum_{i=1}^my_i h_{\bf x}(x, x_i)$
with a measurable function $h_{\bf x}: \mathcal X \times \mathcal X
\to \mathbb R$ that depends on ${\bf x}:=\{x_i\}_{i=1}^m$, then
\begin{equation}\label{unbias}
     \mathbf E\left[\|f_S-f_\rho\|_\mu^2| {\bf x}\right]=\mathbf
     E\left[\left\|f_S-\sum_{i=1}^mf_\rho(x_i)h_{\bf x}(\cdot, x_i)\right\|_\mu^2| {\bf x}\right]+
      \left\|\sum_{i=1}^mf_\rho(x_i)h_{\bf x}(\cdot,
      x_i)-f_\rho\right\|_\mu^2
\end{equation}
for any Borel probability measure $\mu$ on $\mathcal X$.
\end{lemma}

{\bf Proof.} Since $f_\rho(x)$ is the conditional mean of $y$ given
$x\in\mathcal X$, we have from $f_{S}(x)=\sum_{i=1}^my_ih_{{\bf
x}}(x, x_i)$ that $ \mathbf E[f_S|{\bf x}]=
\sum_{i=1}^mf_\rho(x_i)h_{\bf x}(\cdot, x_i)$. Hence,
\begin{eqnarray*}
      &&\mathbf E\left[\left\langle f_S-\sum_{i=1}^mf_\rho(x_i)h_{\bf x}(\cdot, x_i),
       \sum_{i=1}^mf_\rho(x_i)h_{\bf x}(\cdot, x_i) -f_\rho\right\rangle_\mu|{\bf x}\right]\\
       &=&
        \left\langle \mathbf E\left[f_S|{\bf x}\right]-\sum_{i=1}^mf_\rho(x_i)h_{\bf x}(\cdot, x_i),
       \sum_{i=1}^mf_\rho(x_i)h_{\bf x}(\cdot, x_i)-f_\rho\right\rangle_\mu =0.
\end{eqnarray*}
Thus, along with the inner-product expression
\begin{eqnarray*}
       \|f_S-f_\rho\|_\mu^2
       &=&\left\|f_S-\sum_{i=1}^mf_\rho(x_i)h_{\bf x}(\cdot, x_i)\right\|_\mu^2+
       \left\|\sum_{i=1}^mf_\rho(x_i)h_{\bf x}(\cdot, x_i)-f_\rho\right\|_\mu^2\\
       &+&
       2\left\langle f_S-\sum_{i=1}^mf_\rho(x_i)h_{\bf x}(\cdot, x_i),
       \sum_{i=1}^mf_\rho(x_i)h_{\bf x}(\cdot, x_i)-f_\rho\right\rangle_\mu
\end{eqnarray*}
the above equality yields the desired result (\ref{unbias}). This
completes the proof of Lemma \ref{Lemma:important2}. $\Box$

We are now ready to prove the two main results of the paper.

%

{\bf Proof of Theorem \ref{Theorem: optimal rate without feedback}.}
We divide the proof into four steps, namely: error decomposition,
sampling error estimation, approximation error estimation, and
learning rate deduction.

{\it Step 1: Error decomposition.} Let $\dot{H}_{\bf k,j}$ be the
set of interior points of $H_{\bf k,j}$. For arbitrarily fixed ${\bf
k',j'}$ and $x\in \dot{H}_{\bf k',j'}$, it follows from (\ref{N1&3})
that
\begin{eqnarray*}
      \sum_{{\bf j}\in \mathbb N_{2 q^*}^D} \sum_{{\bf k}\in \mathbb N_{2n}^d}
      \sum_{i=1}^m N_{1,D, q^*, \zeta_{{\bf j}}} (x_i)
      N_{3,{\bf k,j}}(x_i) y_i N_{3,{\bf k,j}}(x)
      &=&
      \sum_{i=1}^my_iN_{1,D, q^*, \zeta_{{\bf j'}}} (x_i)
      N_{3,{\bf k',j'}}(x_i)\\
      &=&\sum_{i=1}^my_iI_{H_{\bf k',j'}}(x_i).
\end{eqnarray*}
If, in addition, each $x_i\in \dot{H}_{\bf k,j}$ for some ${\bf
k,j}\in \mathbb N_{2 q^*}^D\times \mathbb N_{2n}^d$, then  we have,
from (\ref{final estimator1}), that
\begin{equation}\label{rewritten N3}
           N_{3}(x)=\frac{\sum_{i=1}^my_iI_{H_{\bf k',j'}}(x_i)}{
           \sum_{i=1}^mI_{H_{\bf k',j'}}(x_i)}
           =
           \frac{\sum_{i=1}^my_iI_{H_{\bf k',j'}}(x_i)}{T_{\bf
           k',j'}(S)}.
\end{equation}
In view of Assumption \ref{Assumption:rhox}, it follows that for an
arbitrary subset $A\subset R^{D}$, $\lambda_G(A)=0$ implies
$\rho_X(A)=0$, where $\lambda_G (A)$ denotes the geodesic metric of
the Riemannian manifold $\mathcal X$. In particular, for $A=H_{\bf
k,j}\backslash\dot{H}_{\bf k,j}$ in the above analysis, we have
$\rho_X(H_{\bf k,j}\backslash\dot{H}_{\bf k,j})=0$, which implies
that (\ref{rewritten N3}) almost surely holds. Next, set
\begin{equation}\label{Definition of tilde n4444}
      \widetilde{N_{3}}  :=\mathbf
       E\left[N_3 |{\bf x}\right].
\end{equation}
Then it follows from Lemma \ref{Lemma:important2}, with
$\mu=\rho_X$, that
\begin{equation}\label{Error decomposition 222}
   \mathbf E\left[\|N_3-f_\rho\|_\rho^2\right]
          =
        \mathbf E\left[\| N_3-\widetilde{N_3}\|_\rho^2\right]
        +
        \mathbf
        E\left[\|\widetilde{N_3}-f_\rho\|_\rho^2\right].
\end{equation}
In what follows, the two terms on the right-hand side of (\ref{Error
decomposition 222}) will be called sampling error and approximation
error, respectively.

{\it Step 2: Sampling error estimation.}   Due to Assumption
\ref{Assumption:rhox}, we have
\begin{eqnarray}\label{sam 2.1}
      \mathbf E[\|N_3 -\widetilde{N_3}\|_\rho^2]
      =
     \sum_{({\bf j,k})\in\mathbb N_{2q^*}^D\times\mathbb N_{2n}^d}
      \int_{\dot{H}_{\bf
     k,j}}
     \mathbf
     E\left[(N_3 (x)-\widetilde{N_3} (x))^2\right]d\rho_X.
\end{eqnarray}
On the other hand, (\ref{rewritten N3}) and (\ref{Definition of
tilde n4444}) together imply that
$$
           N_{3}(x)-\widetilde{N_3}(x)=
           \frac{\sum_{i=1}^m(y_i-f_\rho(x_i))I_{H_{\bf k,j}}(x_i)}{T_{\bf
           k,j}(S)}
$$
almost surely for $x\in \dot{H}_{\bf k,j}$, and that
$$
      \mathbf E\left[(N_3 (x)-\widetilde{N_3} (x))^2|{\bf x}\right]
      =
     \frac{\sum_{i=1}^m\int_{\mathcal Y}(y-f_\rho(x_i))^2d\rho(y|x_i)I^2_{H_{\bf k,j}}(x_i)}{[T_{\bf
           k,j}(S)]^2}
           \leq 4M^2\frac{I_{\{z:T_{\bf k,j}(z)>0\}}(S)}{T_{\bf
           k,j}(S)},
$$
where $\mathbb E[y_i|x_i]=f_\rho(x_i)$ in the second equality,
$I^2_{H_{\bf k,j}}(x_i)=I_{H_{\bf k,j}}(x_i)$ and $|y_i|\leq M$
holds almost surely in the inequality. It then follows from Lemma
\ref{Lemma:important} and Assumption \ref{Assumption:rhox} that
$$
       \mathbf E\left[(N_3 (x)-\widetilde{N_3} (x))^2\right]
       \leq \frac{8 M^2}{(m+1)\rho_X(H_{\bf k, j})}.
$$
This, together with (\ref{sam 2.1}), implies that
\begin{eqnarray}\label{sam 2.est}
      \mathbf E[\|N_3 -\widetilde{N_3}\|_\rho^2]
      \leq
     \sum_{({\bf j,k})\in\mathbb N_{2q^*}^D\times\mathbb N_{2n}^d}
      \int_{\dot{H}_{\bf
     k,j}}
      \frac{8M^2}{(m+1)\rho_X(H_{\bf k, j})}d\rho_X
      \leq
      \frac{8(2q^*)^D(2n)^dM^2}{m+1}.
\end{eqnarray}

{\it Step 3: Approximation error estimation.} According to
Assumption \ref{Assumption:rhox}, we have
\begin{eqnarray}\label{app 2.1}
      \mathbf E[\|f_\rho -\widetilde{N_3}\|_\rho^2]
      =
     \sum_{({\bf j,k})\in\mathbb N_{2q^*}^D\times\mathbb N_{2n}^d}
      \int_{\dot{H}_{\bf
     k,j}}
     \mathbf
     E\left[(f_\rho (x)-\widetilde{N_3} (x))^2\right]d\rho_X.
\end{eqnarray}
For  $x\in \dot{H}_{\bf
     k,j}$, it follows from  Assumption \ref{Assumption:frho}, (\ref{rewritten N3}) and
(\ref{Definition of tilde n4444}) that
\begin{eqnarray*}
      &&\left|f_\rho (x)-\widetilde{N_3} (x)\right|\leq
      \frac{\sum_{i=1}^m|f_\rho(x)-f_\rho(x_i)|I_{H_{\bf k,j}}(x_i)}{T_{\bf
           k,j}(S)}\leq c_0(\max_{x,x'\in H_{\bf k,j}}d_G(x,x'))^s
\end{eqnarray*}
almost surely holds. We then have, from  (\ref{equality of distance
of N2}) and $N_{2,{\bf j}}(x),N_{2,{\bf j}}(x') \in A_{d,1/n,t_{\bf
k}}$, that
$$
      \max_{x,x'\in H_{{\bf k},{\bf j}}} d_G(x,x')
       \leq
      \max_{x,x'\in H_{{\bf k},{\bf j}}} \alpha^{-1}
       \|N_{2,{\bf j}}(x)-N_{2,{\bf j}}(x')\|_d.
$$
Now, since $\max_{t,t'\in A_{d,1/n,t_{\bf k}}}\|t-t'\|_d\leq
\frac{2\sqrt{d}}n$, we obtain
$$
      \max_{x,x'\in H_{{\bf k},{\bf j}}} d_G(x,x')
     \leq  \frac{2 d^{1/2}}{\alpha }n^{-1},
$$
so that
$$
        \left|f_\rho (x)-\widetilde{N_3} (x)\right|\leq
        c_0\frac{2^s d^{s/2}}{\alpha^s}n^{-s}.
$$
almost surely holds. Inserting the above estimate into (\ref{app
2.1}), we also obtain
\begin{eqnarray}\label{app 2.est}
      \mathbf E[\|f_\rho -\widetilde{N_3}\|_\rho^2]
      \leq
     \sum_{({\bf j,k})\in\mathbb N_{2q^*}^D\times\mathbb N_{2n}^d}
      \rho_X({\dot{H}_{\bf
     k,j}})\frac{c_0^24^s d^{s}}{\alpha^{2s}}n^{-2s}
     \leq
     \frac{c_0^24^s d^{s}}{\alpha^{2s}}n^{-2s}.
\end{eqnarray}
{\it Step 4: Learning rate deduction.} Inserting (\ref{app 2.est})
and (\ref{sam 2.est}) into (\ref{Error decomposition 222}), we
obtain
\begin{eqnarray*}
        &&\mathbf E\left[\|N_3 -f_\rho\|_\rho^2\right]
          \leq
       \frac{8(2q^*)^D(2n)^dM^2}{m+1}
        +
       \frac{c_0^24^s d^{s}}{\alpha^{2s}}n^{-2s}.
\end{eqnarray*}
Since $n=\lceil m^{1/(2s+d)}\rceil$, we have
$$
    \mathbf E\left[\|N_3^F-f_\rho\|_\rho^2\right]\leq
    C_1m^{-\frac{2s}{2s+d}}
$$
with
$$
    C_1:=8(2q^*)^D2^dM^2+\frac{c_0^24^{s}d^{s}}{\alpha^{2s}}.
$$
As $q^*$  depends only on $\mathcal X$, $C_2'$ is independent of $m$
or $n$.

This completes the proof of Theorem \ref{Theorem: optimal rate
without feedback}. $\Box$

{\bf Proof of Theorem \ref{Theorem: optimal rate}.} Similar to the
proof of Theorem \ref{Theorem: optimal rate without feedback}, we
also divide this proof into four steps.

 {\it Step 1: Error
decomposition.}  From (\ref{feedback 1}), we have
\begin{equation}\label{defienition of feedback}
      N_3^F(x)= \sum_{i=1}^m y_i h_{\bf x}(x, x_i),
\end{equation}
where $h_{\bf x}: \mathcal X \times \mathcal X \to \mathbb R$ is a
function defined for $x, u \in \mathcal X$ by
\begin{equation}\label{def h}
h_{\bf x}(x, u) = \frac{\sum_{{\bf j}\in \mathbb N_{2 q^*}^D}
\sum_{{\bf k}\in \mathbb N_{2n}^d} \Phi_{\bf k,j}(x, u)}{\sum_{{\bf
j}\in \mathbb N_{2 q^*}^D} \sum_{{\bf k}\in \mathbb N_{2n}^d}
\sum_{i=1}^m \Phi_{\bf k,j}(x, x_i)},
\end{equation}
and $h_{\bf x}(x, u)=0$ when the denominator vanishes.

Define $\widetilde{N^F_{3}}: \mathcal X \to \mathbb R$ by
\begin{equation}\label{Definition of tilde n3}
      \widetilde{N^F_{3}}(x) =\mathbf
       E\left[N^F_3(x)| {\bf x}\right]=\sum_{i=1}^m f_\rho (x_i) h_{\bf x}(x, x_i).
\end{equation}
Then it follows from Lemma \ref{Lemma:important2} with $\mu=\rho_X$,
that
\begin{equation}\label{Error decomposition}
   \mathbf E\left[\|N_3^F-f_\rho\|_\rho^2\right]
          =
        \mathbf E\left[\| N_3^F-\widetilde{N^F_3}\|_\rho^2\right]
        +
        \mathbf
        E\left[\|\widetilde{N^F_3}-f_\rho\|_\rho^2\right].
\end{equation}
In what follows, the terms on the right-hand side of (\ref{Error
decomposition}) will be called sampling error  and approximation
error, respectively.


By (\ref{N1&3}), for each $x\in \mathcal X$ and $i\in\{1, \cdots,
m\}$, we have $\Phi_{\bf k,j}(x, x_i) = I_{H_{\bf k,j}}(x_i)
N_{3,{\bf k,j}}(x) = I_{H_{\bf k, j}}(x_i)$ for $({\bf
j,k})\in\Lambda_x$ and  $\Phi_{\bf k,j}(x, x_i) = 0$ for $({\bf
j,k})\notin\Lambda_x$, where $\Lambda_x$ is defined by
(\ref{Lambdaset}). This, together with (\ref{Definition of tilde
n3}), (\ref{defienition of feedback}) and (\ref{def h}), yields both
\begin{equation}\label{fdiff}
      N_3^F(x) - \widetilde{N^F_{3}}(x)
      = \sum_{i=1}^m \left(y_i - f_\rho (x_i)\right)
         \frac{\sum_{({\bf j}, {\bf k})\in \Lambda_x}
         I_{H_{\bf k, j}}(x_i)}{\sum_{({\bf j}, {\bf k})\in \Lambda_x} T_{{\bf k, j}}(S)},
         \qquad \forall x\in  \mathcal X
\end{equation}
and
\begin{equation}\label{fdiff11}
      \widetilde{N^F_{3}}(x)-f_\rho(x)
      = \sum_{i=1}^m  [f_\rho (x_i)-f_\rho(x)]
         \frac{\sum_{({\bf j}, {\bf k})\in \Lambda_x}
         I_{H_{\bf k, j}}(x_i)}{\sum_{({\bf j}, {\bf k})\in \Lambda_x} T_{{\bf k, j}}(S)},
         \qquad \forall x\in  \mathcal X,
\end{equation}
where $T_{{\bf k, j}}(S)=\sum_{i=1}^mI_{H_{\bf k, j}}(x_i).$

{\it Step 2:  Sampling error estimation.} First consider
\begin{equation}\label{sam 1}
        \mathbf E \left[\|N_3^F-\widetilde{N^F_3}\|_\rho^2\right]
        \leq \sum_{({\bf j,k}) \in \mathbb N_{2q^*}^D
        \times \mathbb N_{2n}^d} \int_{H_{\bf k,j}}
        \mathbf E \left[\left(N^F_3(x)-\widetilde{N_3^F}(x)\right)^2\right] d\rho_X.
\end{equation}
Then for each $x\in H_{\bf k,j}$, since $\mathbb E[y|x]=f_\rho(x)$,
it follows from (\ref{fdiff}) and $|y|\leq M$ that
\begin{eqnarray*}
      &&\mathbf E \left[\left(N^F_3(x)-\widetilde{N_3^F}(x)\right)^2|{\bf x}\right]
      =
      \mathbf E\left[\left(\sum_{i=1}^m \left(y_i - f_\rho (x_i)\right)
         \frac{\sum_{({\bf j}, {\bf k})\in \Lambda_x}
         I_{H_{\bf k, j}}(x_i)}{\sum_{({\bf j}, {\bf k})\in \Lambda_x} T_{{\bf k,
         j}}(S)}\right)^2\big|{\bf x}\right]\\
         &=&
       \mathbf E\left[ \sum_{i=1}^m \left(y_i - f_\rho
       (x_i)\right)^2
         \left(\frac{\sum_{({\bf j}, {\bf k})\in \Lambda_x}
         I_{H_{\bf k, j}}(x_i)}{\sum_{({\bf j}, {\bf k})\in \Lambda_x} T_{{\bf k,
         j}}(S)}\right)^2\big|{\bf x}\right]\\
         &\leq&
         4M^2\sum_{i=1}^m\left(\frac{\sum_{({\bf j}, {\bf k})\in \Lambda_x}
         I_{H_{\bf k, j}}(x_i)}{\sum_{({\bf j}, {\bf k})\in \Lambda_x} T_{{\bf k,
         j}}(S)}\right)^2
\end{eqnarray*}
almost surely holds. Hence, since $\sum_{i=1}^m I_{H_{\bf k,
j}}(x_i)= T_{{\bf k, j}}(S)$, we may apply the Schwarz inequality to
$\sum_{({\bf j}, {\bf k})\in \Lambda_x}I_{H_{\bf k, j}}(x_i)$ to
obtain
\begin{eqnarray*}
      &&\mathbf E \left[\left(N^F_3(x)-\widetilde{N_3^F}(x)\right)^2|{\bf x}\right]
      \leq
      \frac{4M^2|\Lambda_x|\sum_{({\bf j}, {\bf k})\in \Lambda_x}\sum_{i=1}^mI^2_{H_{\bf k, j}}(x_i)}{\left(\sum_{({\bf j}, {\bf k})\in \Lambda_x} T_{{\bf k,
         j}}(S)\right)^2}\\
      &=&
    \frac{4M^2|\Lambda_x|I_{\{z\in\mathcal Z^m:\sum_{({\bf j}, {\bf k})\in \Lambda_x}T_{{\bf k,j}}>0\}} (S)}{\sum_{({\bf j}, {\bf k})\in \Lambda_x} T_{{\bf k,
         j}}(S)}.
\end{eqnarray*}
Thus, it follows from Lemma \ref{Lemma:important} and (\ref{cap1})
that
\begin{eqnarray*}
      \mathbf E \left[\left(N^F_3(x)-\widetilde{N_3^F}(x)\right)^2\right]
      &=&
      \mathbf E\left[\mathbf E \left[\left(N^F_3(x)-\widetilde{N_3^F}(x)\right)^2|{\bf
      x}\right]\right]\\
      &\leq&
      \frac{8M^22^{D+d} }{(m+1)\rho_X(\cup_{({\bf
       j,k})\in\Lambda_x}H_{\bf k, j})}.
\end{eqnarray*}
This, along with (\ref{sam 1}), implies that
\begin{eqnarray}\label{Sam est}
        &&\mathbf E \left[\|N_3^F-\widetilde{N^F_3}\|_\rho^2\right]
         \leq
         \frac{2^{D+d+3}M^2 }{(m+1)} \sum_{({\bf j,k}) \in \mathbb N_{2q^*}^D
        \times \mathbb N_{2n}^d} \int_{H_{\bf k,j}}
         \frac{1}{\rho_X(\cup_{({\bf
       j,k})\in\Lambda_x}H_{\bf k, j})} d\rho_X\nonumber\\
       &\leq&
       \frac{2^{D+d+3}M^2 }{(m+1)} \sum_{({\bf j,k}) \in \mathbb N_{2q^*}^D
        \times \mathbb N_{2n}^d} \int_{H_{\bf k,j}}
         \frac{1}{\rho_X(H_{\bf k, j})} d\rho_X
       \leq
       \frac{2^{D+d+3}{(2q^*)}^DM^2{(2n)}^d }{(m+1)}.
\end{eqnarray}

{\it Step 3: Approximation error estimation.} For each $x\in
\mathcal X$, set
$$
     A_1(x):=
      \mathbf E\left[( \widetilde{N_3^F}(x)-f_\rho(x))^2| \sum_{({\bf j}, {\bf k})\in \Lambda_x}T_{{\bf k,j}}(S) =0\right]
      Pr\left[ \sum_{({\bf j}, {\bf k})\in \Lambda_x}T_{{\bf k,j}}(S)
      =0\right]
$$
and
$$
     A_2(x):=
     \mathbf E\left[( \widetilde{N_3^F}(x)-f_\rho(x))^2| \sum_{({\bf j}, {\bf k})\in \Lambda_x}T_{{\bf k,j}}(S) \geq1\right]
     Pr\left[\sum_{({\bf j}, {\bf k})\in \Lambda_x}T_{{\bf k,j}}(S)\geq
     1\right];
$$
and observe that
\begin{equation}\label{app 1}
        \mathbf E \left[\| \widetilde{N^F_3}-f_\rho\|_\rho^2\right]
       =
      \int_{\mathcal X}
        \mathbf E \left[\left( \widetilde{N_3^F}(x)-f_\rho(x)\right)^2\right] d\rho_X
        = \int_{\mathcal X}A_1(x)d\rho_X+\int_{\mathcal X}A_2(x)d\rho_X.
\end{equation}
Let us first consider $\int_{\mathcal X}A_1(x)d\rho_X$ as follows.
Since $\widetilde{N_3^F}(x)=0$ for $\sum_{({\bf j}, {\bf k})\in
\Lambda_x}T_{{\bf k,j}}(S) =0$, we have, from $
      |f_\rho(x)|\leq M$,  that
$$
   \mathbf E\left[(
      \widetilde{N_3^F(x)}-f_\rho(x))^2| \sum_{({\bf j}, {\bf k})\in \Lambda_x}T_{{\bf k,j}}(S)=0\right]\leq
      M^2.
$$
On the other hand, since
$$
      Pr\left[\sum_{({\bf j}, {\bf k})\in \Lambda_x}T_{{\bf k,j}}(S)=0\right]
      =[1-\rho_X(\cup_{({\bf j,k})\in \Lambda_x}H_{\bf k,j })]^{m},
$$
it follows from the elementary inequality
$$
     v(1-v)^m\leq ve^{-mv}\leq\frac1{em},\qquad\forall 0\leq v\leq 1
$$
that
\begin{eqnarray}\label{Bound A_1(x)}
    &&\int_{\mathcal X}A_1(x)d\rho_X
    \leq
     \int_{\mathcal X}M^2[1-\rho_X(\cup_{({\bf j,k})\in \Lambda_x}H_{\bf k,j
     })]^{m}d\rho_X \nonumber\\
     &\leq&
     M^2\sum_{({\bf j',k'})\in\mathbb N_{2q^*}^D\times\mathbb N_{2n}^d}\int_{{H}_{\bf
     k',j'}}[1-\rho_X(\cup_{({\bf j,k})\in \Lambda_x}H_{\bf k,j
     })]^{m}d\rho_X \nonumber\\
      &\leq&
    M^2\sum_{({\bf j,k})\in\mathbb N_{2q^*}^D\times\mathbb N_{2n}^d}
      \int_{{H}_{\bf
     k,j}}[1-\rho_X(H_{\bf k,j })]^{m}d\rho_X
    \leq
    M^2\sum_{({\bf j,k})\in\mathbb N_{2q^*}^D\times\mathbb N_{2n}^d}[1-\rho_X(H_{\bf k,j
    })]^{m}\rho_X({H}_{\bf
     k,j})\nonumber\\
     &\leq&
     \frac{(2n)^d(2q^*)^DM^2}{em}.
\end{eqnarray}

We next consider  $\int_{\mathcal X}A_2(x)d\rho_X$. Let  $x\in=
\mathcal X$ be so chosen that  $\sum_{({\bf j,k})\in
\Lambda_x}T_{{\bf k,j}}(S)\geq 1$. Then $x_i\in H_x:=\cup_{({\bf
j,k})\in\Lambda_x}H_{\bf k,j}$ at least for some
$i\in\{1,2,\dots,m\}$. For those $x_i\notin H_x$, we have
$\sum_{({\bf j,k})\in \Lambda_x}I_{H_{\bf k,j}}(x_i)=0$, so that
\begin{eqnarray*}
      &&\left|\widetilde{N^F_{3}}(x)-f_\rho(x)\right|
     = \sum_{i:x_i\in H_x}  |f_\rho (x_i)-f_\rho(x)|
         \frac{\sum_{({\bf j}, {\bf k})\in \Lambda_x}
         I_{H_{\bf k, j}}(x_i)}{\sum_{({\bf j}, {\bf k})\in \Lambda_x} T_{{\bf k,
         j}}(S)}.
\end{eqnarray*}
For $x_i\in H_x$, we have $x_i\in H_{\bf k,j}$ for some $({\bf
j,k})\in\Lambda_x$. But $x\in H_{\bf k,j}$, so that
\begin{eqnarray*}
      |\widetilde{N_3^F}(x)  -f_\rho(x)|
       \leq \max_{u,u'\in  H_{{\bf k},{\bf
       j}}}|f_\rho(u)-f_\rho(u')\leq c_0 \max_{u,u'\in  H_{{\bf k},{\bf
       j}}}[d_G(u,u')]^s,\qquad x\in\mathcal X.
\end{eqnarray*}
But  (\ref{equality of distance of N2}) implies that

\begin{eqnarray*}
      \max_{u,u'\in { H_{{\bf k},{\bf j}}}} [d_G(u,u')]^s
       &\leq&
      \max_{u,u'\in { H_{{\bf k},{\bf j}}}}\alpha^{-s}
       \|N_{2,{\bf j}_x}(u)-N_{2,{\bf j}_x}(u')\|^s_d
       \leq
       \alpha^{-s}\max_{t,t'\in A_{d,1/n,t_{{\bf
       k}}}}\|t-t'\|^s_d\\
       &\leq&
       \frac{2^sd^{s/2}}{\alpha^s}n^{-s}.
\end{eqnarray*}
Hence, for $x\in\mathcal X$ with $\sum_{({\bf
j,k})\in\Lambda_x}T_{\bf k,j}(S)\geq 1$, we have
$$
     |\widetilde{N_3^F}(x)  -f_\rho(x)|
     \leq  \frac{c_02^{s}d^{s/2}}{\alpha^s}n^{-s}\frac{\sum_{i:x_i\in
     H_x}
     \sum_{({\bf j,k})\in\Lambda_x}}{\sum_{({\bf j,k})\in\Lambda_x}T_{\bf
     k,j}(S)}
     \leq  \frac{c_02^{s}d^{s/2}}{\alpha^s}n^{-s},\qquad \forall x\in
     \mathcal X,
$$
or equivalently,
\begin{equation}\label{Bound A_2}
   \int_{\mathcal X}A_2(x)d\rho_X
   \leq \int_{\mathcal X}\mathbf E\left[( \widetilde{N_3^F}(x)-f_\rho(x))^2|
    \sum_{({\bf j},
      {\bf k})\in \Lambda_x}T_{{\bf k,j}}(S) \geq1\right]d\rho_X
     \leq
     \frac{c_0^24^{s}d^{s}}{\alpha^{2s}}n^{-2s}.
\end{equation}
Therefore, putting (\ref{Bound A_1(x)}) and (\ref{Bound A_2}) into
(\ref{app 1}), we have
\begin{equation}\label{app est}
       \mathbf E \left[\| \widetilde{N^F_3}-f_\rho\|_\rho^2\right]
       \leq
     \frac{c_0^24^{s}d^{s}}{\alpha^{2s}}n^{-2s}+\frac{M^2(2n)^d(2q^*)^D}{em}.
\end{equation}

{\it Step 4: Learning rate deduction.} By inserting (\ref{Sam est})
and (\ref{app est}) into (\ref{Error decomposition}), we obtain
\begin{eqnarray*}
        &&\mathbf E\left[\|N_3^F-f_\rho\|_\rho^2\right]
          \leq
      \frac{2^{D+d+3}{(2q^*)}^DM^2{(2n)}^d }{m+1}
        +
      \frac{c_0^24^{s}d^{s}}{\alpha^{2s}}n^{-2s}+\frac{M^2(2n)^d(2q^*)^D}{em}.
\end{eqnarray*}
Hence, in view of $n=\lceil m^{1/(2s+d)}\rceil$, we have
$$
    \mathbf E\left[\|N_3^F-f_\rho\|_\rho^2\right]\leq
    C_2m^{-\frac{2s}{2s+d}}
$$
with
$$
    C_2:= 2^{D+d+4}{(2q^*)}^DM^2{(2n)}^d +\frac{c_0^24^{s}d^{s}}{\alpha^{2s}}.
$$
This completes the proof of Theorem \ref{Theorem: optimal rate},
since $q^*$  depends only on $\mathcal X$, so that $C_2'$ is
independent of $m$ or $n$. $\Box$

\section*{Acknowledgement}
 The research of C. K. Chui is partially supported by U.S. ARO Grant
W911NF-15-1-0385, Hong Kong Research Council (Grant No. 12300917),
and Hong Kong Baptist University (Grant No. HKBU-RC-ICRS/16-17/03).
The research of S. B. Lin is partially supported by the National
Natural Science Foundation of China (Grant No. 61502342). The work
of D. X. Zhou is supported partially by the Research Grants Council
of Hong Kong [Project No. CityU 11303915] and by National Natural
Science Foundation of China under Grant 11461161006. Part of the
work was done during a visit to Shanghai Jiaotong University (SJTU),
for which the support from SJTU and Ministry of Education is greatly
appreciated.

\end{document}